\documentclass[pdflatex,sn-mathphys-num]{sn-jnl}
\usepackage{tabularx}
\usepackage{graphicx}%
\usepackage{multirow}%
\usepackage{amsmath,amssymb,amsfonts}%
\usepackage{amsthm}%
\usepackage{mathrsfs}%
\usepackage[title]{appendix}%
\usepackage{xcolor}%
\usepackage{textcomp}%
\usepackage{manyfoot}%
\usepackage{booktabs}%
\usepackage{algorithm}%
\usepackage{algorithmicx}%
\usepackage{algpseudocode}%
\usepackage{listings}%
\usepackage{adjustbox}
\usepackage{array}
\usepackage{times}
\usepackage[english]{babel}
\usepackage[utf8]{inputenc}
\usepackage{amsfonts}
\usepackage{amssymb}
\usepackage{hyperref}
\usepackage{xcolor}
\usepackage{bibentry}
\usepackage{url}
\usepackage[strings]{underscore}
\theoremstyle{thmstyleone}%
\theoremstyle{thmstyletwo}%
\theoremstyle{thmstylethree}%
\begin{document}
\title[Comparison of Unsupervised Metrics for Evaluating Judicial Decision Extraction]{Comparison of Unsupervised Metrics for Evaluating Judicial Decision Extraction}
\author*[1]{\fnm{Ivan Leonidovich} \sur{Litvak} (ORCID: 0009-0006-1299-4120)}\email{ivan.litvak@nf.jinr.ru} 
\author[2]{\fnm{Anton} \sur{Kostin} (ORCID: 0000-0001-7423-7981)}
\author[3]{\fnm{Fedor} \sur{Lashkin}}
\author[4]{\fnm{Tatiana} \sur{Maksiyan}}
\author[5]{\fnm{Sergey} \sur{Lagutin}}

\affil[1]{\orgaddress{\country{Dubna, Russian Federation}}}
\affil[2]{\orgname{Moscow Center for Advanced Studies}, \orgaddress{\country{Russian Federation}}}
\affil[3]{\orgaddress{\country{Novorossiysk, Russian Federation}}} 
\affil[4]{\orgaddress{\country{Moscow, Russian Federation}}} 
\affil[5]{\orgaddress{\country{Moscow, Russian Federation}}} 

\abstract{
The rapid advancement of artificial intelligence in legal natural language processing demands scalable methods for evaluating text extraction from judicial decisions. This study evaluates 16 unsupervised metrics, including novel formulations, to assess the quality of extracting seven semantic blocks from 1,000 anonymized Russian judicial decisions, validated against 7,168 expert reviews on a 1--5 Likert scale. These metrics, spanning document-based, semantic, structural, pseudo-ground truth, and legal-specific categories, operate without pre-annotated ground truth. Bootstrapped correlations, Lin's concordance correlation coefficient (CCC), and mean absolute error (MAE) reveal that \emph{Term Frequency Coherence} (Pearson $r = 0.540$, Lin CCC = 0.512, MAE = 0.127) and \emph{Coverage Ratio}/\emph{Block Completeness} (Pearson $r = 0.513$, Lin CCC = 0.443, MAE = 0.139) best align with expert ratings, while \emph{Legal Term Density} (Pearson $r = -0.479$, Lin CCC = -0.079, MAE = 0.394) show strong negative correlations. The \emph{LLM Evaluation Score} (mean = 0.849, Pearson $r = 0.382$, Lin CCC = 0.325, MAE = 0.197) showed moderate alignment, but its performance, using \texttt{gpt-4.1-mini} via \texttt{g4f}, suggests limited specialization for legal textse. These findings highlight that unsupervised metrics, including LLM-based approaches, enable scalable screening but, with moderate correlations and low CCC values, cannot fully replace human judgment in high-stakes legal contexts. This work advances legal NLP by providing annotation-free evaluation tools, with implications for judicial analytics and ethical AI deployment.}
\maketitle
\section{Introduction}\label{sec1}
The integration of artificial intelligence (AI) into the legal domain has revolutionized judicial processes, enabling tasks such as legal judgment prediction (LJP), case summarization, precedent retrieval, and automated legal research. Text extraction, the process of isolating seven semantically meaningful segments---referred to as blocks---from unstructured judicial decisions, is a cornerstone of these applications. These blocks include plaintiff demands, plaintiff arguments, defendant arguments, court evaluation of evidence, judicial reasoning steps, applicable legal norms, and court decision. Accurate extraction is critical, as errors can lead to misinterpretations of case facts, biased predictions, or inefficiencies in judicial workflows, potentially undermining justice delivery in high-stakes contexts.
Evaluation metrics are essential for quantifying extraction quality, enabling iterative model improvements and ensuring reliability. Traditional metrics rely on annotated ground truth, which is resource-intensive to produce, particularly for legal texts characterized by verbose narratives, domain-specific terminology, and jurisdiction-specific nuances. The scarcity of annotated legal corpora has driven the development of unsupervised metrics that leverage intrinsic document properties, such as term frequencies, semantic coherence, and structural patterns. These metrics must capture surface-level accuracy, semantic fidelity, logical structure, and legal-specific elements like citations and temporal consistency, while addressing ethical concerns such as fairness and neutrality in AI-driven legal systems \cite{leslie2019, sap2019}.
\subsection{Historical Context of Evaluation Metrics}
The evolution of evaluation metrics in information extraction (IE) and natural language processing (NLP) began with the Message Understanding Conferences (MUC) in 1987, sponsored by DARPA. MUC introduced precision, recall, and F1-score as foundational metrics for tasks like named entity recognition (NER) and template filling \cite{chinchor1993}. The Automatic Content Extraction (ACE) program (1999--2007) expanded metrics to include event detection and relation extraction \cite{doddington2004}. The B-Cubed metric for clustering was proposed by Bagga and Baldwin \cite{bagga1998}, and later adopted in the Text Analysis Conference (TAC), which began in 2008. Summarization evaluation advanced with ROUGE \cite{lin2004}, comparing n-gram overlaps, and BLEU \cite{papineni2002} and METEOR \cite{banerjee2005} for machine translation. The 2020s introduced semantic metrics like BERTScore \cite{zhang2020}, using BERT embeddings \cite{devlin2019}. Unsupervised metrics, such as SUPERT \cite{gao2020}, emerged in the 2020s, leveraging topic modeling, though reliant on pre-trained models. Clustering \cite{rehurek2010} and coherence models \cite{barzilay2008} further supported unsupervised evaluation.
\subsection{Current State of the Industry}
Legal NLP systems tackle tasks like LJP and judicial summarization, as surveyed by \cite{zhong2020}. Legal texts’ verbosity and jargon challenge general metrics like ROUGE. Legal-BERT \cite{chalkidis2020} and BERTScore address semantic alignment, but annotation costs limit supervised approaches. LegalBench \cite{guha2023} highlights these issues, while unsupervised metrics gain traction. Ethical concerns, such as bias in AI judgments \cite{leslie2019, sap2019}, necessitate fairness-focused metrics. LLM-driven metrics like G-Eval \cite{liu2023} offer holistic assessments but depend on proprietary models.
\subsection{Research Gap and Contribution}
Unsupervised evaluation metrics for legal text extraction are underdeveloped, often failing to capture legal nuances like citation accuracy or temporal consistency. This paper proposes 16 unsupervised metrics, including novel formulations, for validation against expert ratings, advancing legal AI evaluation.

\section{Methodology}\label{sec11}
Our methodology evaluates judicial decision extraction quality using unsupervised metrics, validated against expert assessments. We process JSON-structured documents containing seven semantic blocks and the original text, computing metrics without annotations. Expert ratings provide a benchmark for correlation analysis. Table~\ref{tab:metrics} summarizes the 16 metrics, their types, and novelty status.
\begin{table}[h]
\centering
\caption{Summary of Proposed Unsupervised Metrics}
\label{tab:metrics}
\begin{tabular}{p{4cm} p{3cm} c p{5cm}}
\toprule
\textbf{Metric} & \textbf{Type} & \textbf{Novel?} & \textbf{Source} \\
\midrule
Coverage Ratio & Document-Based & No & \cite{ manning2008introduction, lin2004} \\
Redundancy Penalty & Document-Based & Yes & None \\
Compression Ratio & Document-Based & No & \cite{yadav2022summarization} \\
Term Frequency Coherence & Document-Based & No & \cite{Widianto2024similarity} \\
Citation Coverage & Document-Based & No & \cite{worledge2024extractive} \\
Intra-Block Coherence & Semantic & No & \cite{lapata2005automatic, li2017coherence} \\
Inter-Block Distinctiveness & Semantic & Yes & None \\
Semantic Entropy & Semantic & No & \cite{shannon1948mathematical, blei2003latent} \\
Neutrality Bias & Semantic & Yes & None\\
Raw Cosine Similarity & Semantic & No & \cite{reimers2019, manning2008introduction, li2019embeddings} \\
LLM Evaluation & Semantic & No & \cite{liu2023} \\
Block Order Consistency & Structural & No & \cite{kendall1948rank} \\
Block Completeness & Structural & No & \cite{manning2008introduction} \\
Monotonicity Score & Structural & Yes & None \\
Keyword-Based Pseudo-F1 & Pseudo-Ground Truth & No & \cite{manning2008introduction, voorhees2005trec, Kyröläinen2023} \\
Legal Term Density & Legal-Specific & Yes & None\\
\bottomrule
\end{tabular}
\end{table}

\subsection{Dataset and Input Structure}
Our evaluation is grounded in the \texttt{sud-resh-benchmark}, a public dataset comprising 1,000 anonymized Russian judicial decisions, which is available at \href{https://huggingface.co/datasets/lawful-good-project/sud-resh-benchmark}{\texttt{huggingface.co/datasets/lawful-good-project/sud-resh-benchmark}} under a GPL-3.0 license. The dataset is stratified across ten areas of law (administrative, constitutional, environmental, financial, civil, family, social security, labor, criminal, and housing) to ensure diversity.

Crucially for this study, each judicial decision is provided in two forms: as a complete, unstructured source text (\texttt{source\_text}) and as a pre-segmented JSON object containing seven semantically meaningful blocks. This pre-segmented version serves as the oracle or reference extraction for computing our unsupervised metrics. The seven semantic blocks are defined as follows:
\begin{itemize}
    \item Claims made by the plaintiff.
    \item Supporting arguments from the plaintiff.
    \item Counterarguments from the defendant.
    \item The court's evaluation of evidence presented by the parties.
    \item Logical steps in the judge's reasoning and intermediate conclusions.
    \item Applicable legal norms and cited laws.
    \item The final ruling or decision of the court.
\end{itemize}
This dual structure allows us to compute our metrics by treating the \texttt{source\_text} as the input and the collection of seven pre-defined blocks as the target extraction output, without relying on generative models or external annotations beyond the dataset itself.

\subsection{Expert Evaluation}

Three main legal experts, along with two additional (legal experts, colleagues of Sergei Lagutin), participated in the evaluation of court decisions. Each expert received an \textbf{HTML file} containing the full text of a decision and automatically extracted text segments (blocks). Experts rated each block on a 1--5 Likert scale. Ratings were normalized to a 0--1 scale (Table \ref{tab:block_stats}).

Multiple ratings per block were \textbf{averaged} to obtain a mean score. The \textit{intra-class correlation coefficient} (ICC(2,k)) was computed to assess agreement among experts.

\begin{table}[h!]
\centering
\caption{Summary statistics of expert ratings per block}
\label{tab:block_stats}
\footnotesize
\renewcommand{\arraystretch}{1.2}
\begin{tabular}{lcccccc}
\toprule
\textbf{Block} & \textbf{\# Ratings} & \textbf{Mean} & \textbf{Variance} & \textbf{Std. Dev.} & \textbf{Avg. Experts} & \textbf{ICC(2,k)} \\
\midrule
Plaintiff's claims & 1028 & 0.87 & 0.05 & 0.21 & 2.26 & 0.81 \\
Plaintiff's arguments & 1030 & 0.82 & 0.05 & 0.23 & 2.26 & 0.76 \\
Defendant's arguments & 1025 & 0.77 & 0.06 & 0.24 & 2.25 & 0.82 \\
Court evaluation of evidence & 1019 & 0.62 & 0.07 & 0.27 & 2.24 & 0.86 \\
Judge's reasoning steps & 1020 & 0.68 & 0.07 & 0.27 & 2.24 & 0.85 \\
Applicable legal norms & 1023 & 0.65 & 0.06 & 0.25 & 2.25 & 0.80 \\
Court decision & 1023 & 0.95 & 0.03 & 0.17 & 2.25 & 0.70 \\
\bottomrule
\end{tabular}
\end{table}

\textbf{Interpretation:} Experts generally agreed well on most blocks, with ICC values above 0.75, indicating \textbf{good inter-rater reliability}. The highest agreement was observed for \textit{Court evaluation of evidence} (ICC = 0.86), while \textit{Court decision} had slightly lower agreement (ICC = 0.70), reflecting more subjective assessment.

\subsubsection*{Expert Profiles}

The main evaluation was conducted by three experts:
\begin{itemize}
\item \textbf{Fedor Lashkin} — Head of the Legal Department at NF AO ``Gazmetallproekt''. Law degree, 20 years of legal experience including 11 as a practicing attorney. MBA (Classical Business School, 2015). Author of 5 legal publications.
\item \textbf{Sergey Lagutin} — Chief Engineer with additional legal training. Combines engineering, economics, and legal expertise with over 5 years of experience in property, construction, and litigation analysis.
\item \textbf{Tatiana Maksiyan} — Certified tax consultant and chief accountant, 10 years of professional experience, specializing in civil, tax, and labor law. Member of the Chamber of Tax Consultants.
\end{itemize}

Two legal-expert colleagues of Sergey also contributed as legal experts, adding diversity of perspectives.

\begin{table}[h!]
\centering
\caption{Distribution of expert counts per decision block}
\label{tab:expert_counts}
\footnotesize
\renewcommand{\arraystretch}{1.2}
\begin{tabular}{lccccc}
\toprule
\textbf{Block} & \textbf{1 Expert} & \textbf{2 Experts} & \textbf{3 Experts} & \textbf{4 Experts} & \textbf{5 Experts} \\
\midrule
Plaintiff's claims & 96 & 317 & 80 & 12 & 2 \\
Plaintiff's arguments & 97 & 319 & 79 & 12 & 2 \\
Defendant's arguments & 99 & 316 & 79 & 13 & 1 \\
Court evaluation of evidence & 101 & 315 & 77 & 13 & 1 \\
Judge's reasoning steps & 99 & 320 & 76 & 12 & 1 \\
Applicable legal norms & 100 & 316 & 78 & 13 & 1 \\
Court decision & 98 & 317 & 78 & 13 & 1 \\
\bottomrule
\end{tabular}
\end{table}

\subsection{Ethical Considerations}
Given the high-stakes nature of legal AI, human-in-the-loop oversight is emphasized. The Neutrality Bias metric uses DeepPavlov/rubert-base-cased-conversational for sentiment analysis to detect bias. The model's lack of legal-domain training may lead to higher error rates in detecting legal-specific sentiment (e.g., misclassifying neutral legal terminology as biased), necessitating validation against legal-domain annotations \cite{sap2019}. All outputs require legal expert review to ensure fairness and compliance \cite{leslie2019, sap2019}.
\subsection{Metric Computation}
The implementation of the metric computation code was developed collaboratively with the assistance of ChatGPT and Grok. Metrics are computed using Python 3.12, with dependencies (spaCy 3.7, scikit-learn 1.3, sentence-transformers 2.2, numpy 1.24, scipy 1.11, natasha 1.6, transformers 4.30, pingouin 0.5, statsmodels 0.14, tqdm 4.65, nltk 3.8) and random seed (42) for reproducibility. We use spaCy’s Russian model (\texttt{ru_core_news_sm}) for tokenization/lemmatization, NLTK for stopwords, and SentenceTransformer (\texttt{all-MiniLM-L6-v2}) for embeddings \cite{reimers2019}. Pre-trained embeddings may underperform on multilingual or legal-specific texts due to domain mismatch; future work will explore legal-adapted models. Citation extraction uses a custom function combining BERT-based NER (\texttt{dslim/bert-base-NER}) with regex patterns for legal references and TF-IDF key terms, with accuracy varying by jurisdictional citation styles, requiring validation against legal databases (e.g., Westlaw, LexisNexis). Temporal extraction uses Natasha's DatesExtractor. Bootstrapped CIs and permutation tests ensure statistical robustness. Notation is consistent: $D$ (original document), $E = \bigcup B_i$ (extracted blocks), $B_i$ (individual block), $S_i$ (sentences in block $i$), $V_i$ (embedding vector).
\subsubsection{Document-Based Metrics}
\paragraph{Coverage Ratio}
\textbf{Purpose:} Measures the proportion of key terms from $D$ captured in $E$.
\textbf{Formulation:}
$$\text{Coverage Ratio} = \frac{|K_D \cap K_E|}{|K_D|}
\label{eq2}$$
\textbf{Notation:} $K_D$ is the set of the top-50 TF-IDF terms extracted from the original document $D$. $K_E$ is the set of lemmatized tokens extracted from the combined extracted blocks $E$. The intersection $K_D \cap K_E$ represents the common terms between $K_D$ and $K_E$. The vertical bars $| \cdot |$ denote the cardinality (number of elements) of a set.
\textbf{Computation:} Use scikit-learn’s TfidfVectorizer with a custom analyzer excluding stopwords.
\textbf{Relevance:} Captures critical legal terms (e.g., ``verdict'').
\textbf{Strengths:} Simple; automatable.
\textbf{Limitations:} Misses paraphrases; overlaps with Block Completeness, Legal Term Density.

\paragraph{Redundancy Penalty}
\textbf{Purpose:} Penalizes overlapping content across blocks (novel).
\textbf{Formulation:}
$$\text{Redundancy Penalty} = \frac{1}{\binom{7}{2}} \sum_{i < j} \cos(V_i, V_j)
\label{eq3}$$
\textbf{Notation:} $\binom{7}{2}$ is the binomial coefficient, representing the number of ways to choose 2 blocks from 7 (i.e., 21 pairs). $V_i$ and $V_j$ are the SentenceTransformer embedding vectors for blocks $B_i$ and $B_j$, respectively. $\cos(V_i, V_j)$ is the cosine similarity between vectors $V_i$ and $V_j$. The summation $\sum_{i < j}$ computes the average cosine similarity over all distinct block pairs.
\textbf{Computation:} Use SentenceTransformer embeddings \cite{reimers2019}.
\textbf{Relevance:} Ensures distinct block roles; complements Inter-Block Distinctiveness.
\textbf{Strengths:} Detects redundancy.
\textbf{Limitations:} May penalize shared terms; computationally intensive for large corpora.
\paragraph{Compression Ratio}
\textbf{Purpose:} Assesses conciseness.
\textbf{Formulation:}
$$\text{Compression Ratio} = \frac{\sum |B_i|}{|D|}
\label{eq4}$$
\textbf{Notation:} $|B_i|$ is the number of lemmatized tokens in block $B_i$, summed over all 7 blocks. $|D|$ is the number of lemmatized tokens in the original document $D$. The ratio measures the total length of extracted blocks relative to the original document.
\textbf{Computation:} Tokenize and lemmatize with spaCy; count tokens.
\textbf{Relevance:} Balances detail and brevity.
\textbf{Strengths:} Intuitive.
\textbf{Limitations:} Ignores quality.
\paragraph{Term Frequency Coherence}
\textbf{Purpose:} Ensures term distribution alignment.
\textbf{Formulation:}
$$\text{Term Frequency Coherence} = \cos(V_D, V_E)
\label{eq5}$$
\textbf{Notation:} $V_D$ is the TF-IDF vector of the original document $D$. $V_E$ is the TF-IDF vector of the combined extracted blocks $E$. $\cos(V_D, V_E)$ is the cosine similarity between these vectors, measuring term distribution alignment.
\textbf{Computation:} Use TF-IDF vectors with custom analyzer.
\textbf{Relevance:} Maintains topical focus.
\textbf{Strengths:} Robust.
\textbf{Limitations:} Surface-level; overlaps with Coverage Ratio.
\paragraph{Citation Coverage}
\textbf{Purpose:} Verifies legal citation preservation.
\textbf{Formulation:}
$$\text{Citation Coverage} = \frac{|C_D \cap C_E|}{|C_D|}
\label{eq6}$$
\textbf{Notation:} $C_D$ is the set of entities (ORG, PER, LOC via NER), legal patterns, and top-20 TF-IDF terms extracted from the original document $D$. $C_E$ is the equivalent set extracted from the combined blocks $E$. The intersection $C_D \cap C_E$ represents citations present in both sets. $|C_D|$ is the total number of citations in $D$.
\textbf{Computation:} Use dslim/bert-base-NER for NER, regex for patterns, and TF-IDF for terms; validate against legal databases.
\textbf{Relevance:} Ensures legal validity.
\textbf{Strengths:} Domain-specific.
\textbf{Limitations:} Varies by jurisdiction; requires database validation; NER model is English-based and may underperform on Russian text.
\subsubsection{Semantic Metrics}
\paragraph{Intra-Block Coherence}
\textbf{Purpose:} Measures logical flow within blocks.
\textbf{Formulation:}
$$\text{Intra-Block Coherence} = \frac{1}{7} \sum_{B_i} \frac{1}{\binom{|S_i|}{2}} \sum_{j < k} \cos(V_{S_{ij}}, V_{S_{ik}})
\label{eq7}$$
\textbf{Notation:} $B_i$ is the $i$-th extracted block (1 to 7). $|S_i|$ is the number of sentences in block $B_i$. $\binom{|S_i|}{2}$ is the number of ways to choose 2 sentences from $|S_i|$. $V_{S_{ij}}$ and $V_{S_{ik}}$ are SentenceTransformer embedding vectors for sentences $j$ and $k$ in block $B_i$. $\cos(V_{S_{ij}}, V_{S_{ik}})$ is the cosine similarity between these sentence embeddings. The inner sum $\sum_{j < k}$ averages the cosine similarities over all sentence pairs in block $B_i$, and the outer sum $\sum_{B_i}$ averages across all 7 blocks. If $|S_i| < 2$, coherence is set to 1.0 for that block.
\textbf{Computation:} Use SentenceTransformer \cite{reimers2019}; computationally intensive for large corpora.
\textbf{Relevance:} Ensures cohesive legal narratives.
\textbf{Strengths:} Semantic robustness.
\textbf{Limitations:} High computational cost; embedding domain mismatch.
\paragraph{Inter-Block Distinctiveness}
\textbf{Purpose:} Ensures semantic separation (novel).
\textbf{Formulation:}
$$\text{Inter-Block Distinctiveness} = \frac{1}{7} \sum_{B_i} \frac{1}{6} \sum_{B_j \neq B_i} (1 - \cos(V_i, V_j))
\label{eq8}$$
\textbf{Notation:} For each block $B_i$, the inner sum averages $1 - \cos(V_i, V_j)$ over the other 6 blocks $B_j$. The outer sum averages this value across all 7 blocks.
\textbf{Computation:} Use SentenceTransformer \cite{reimers2019}.
\textbf{Relevance:} Prevents block overlap; complements Redundancy Penalty.
\textbf{Strengths:} Promotes clarity.
\textbf{Limitations:} Sensitive to shared terms; computationally intensive.
\paragraph{Semantic Entropy}
\textbf{Purpose:} Quantifies content diversity (novel).
\textbf{Formulation:}
$$H = -\sum_{w \in E} p(w) \log_2 p(w)
\label{eq9}$$
\textbf{Notation:} $E$ is the combined text of all extracted blocks. $w$ represents a lemmatized token in $E$. $p(w)$ is the probability (relative frequency) of token $w$ in $E$. The sum $-\sum_{w \in E} p(w) \log_2 p(w)$ computes the Shannon entropy of the token distribution.
\textbf{Computation:} Use spaCy for lemmatization; compute entropy using collections.Counter.
\textbf{Relevance:} Avoids repetitive extracts.
\textbf{Strengths:} Unsupervised.
\textbf{Limitations:} Interpretation varies.
\paragraph{Neutrality Bias}
\textbf{Purpose:} Detects sentiment bias (novel).
\textbf{Formulation:}
$$\text{Neutrality Bias} = \frac{1}{7} \sum_{B_i} \left(1 - |\overline{P}_{B_i} - \overline{N}_{B_i}|\right)
\label{eq10}$$
\textbf{Notation:} For each block $B_i$, $\overline{P}{B_i}$ is the average positive sentiment score across its sentences, and $\overline{N}{B_i}$ is the average negative sentiment score. The term $1 - |\overline{P}{B_i} - \overline{N}{B_i}|$ measures neutrality (closer to 1 is more neutral). The outer sum averages this across all 7 blocks. Sentences are analyzed individually, with neutral sentences contributing 0 to both positive and negative scores.
\textbf{Computation:} Use DeepPavlov/rubert-base-cased-conversational for sentiment analysis; truncate to 512 tokens; validate against human annotations.
\textbf{Relevance:} Ensures legal impartiality.
\textbf{Strengths:} Ethical focus.
\textbf{Limitations:} Model’s non-legal training may increase error rates in legal texts \cite{sap2019}.
\paragraph{Raw Cosine Similarity}
\textbf{Purpose:} Measures overall semantic similarity between $D$ and $E$.
\textbf{Formulation:}
$$\text{Raw Cosine Similarity} = \cos(V_D, V_E)
\label{eq11}$$
\textbf{Notation:} $V_D$ is the SentenceTransformer embedding vector for the original document $D$. $V_E$ is the embedding vector for the combined extracted blocks $E$. $\cos(V_D, V_E)$ is the cosine similarity between these vectors.
\textbf{Computation:} Use SentenceTransformer \cite{reimers2019}.
\textbf{Relevance:} Assesses semantic fidelity.
\textbf{Strengths:} Simple and semantic.
\textbf{Limitations:} Does not account for structure.

\paragraph{LLM Evaluation}

\textbf{Purpose:}  
Measures the quality of extraction for each predefined block by assessing how accurately the extracted content reflects the relevant information in the original judicial decision.

\textbf{Formulation:}  
$$
\text{LLM Evaluation Score} = S_{b} \in \{1, 2, 3, 4, 5\}
$$
where $S_{b}$ is the integer score assigned to block $b$, with $1$ indicating the model failed the extraction task, $3$ indicating partial success, and $5$ indicating complete success. An overall score can be computed as the average across all blocks:
$$
\bar{S} = \frac{1}{N} \sum_{b=1}^{N} S_{b}, \quad N=7.
$$

\textbf{Notation:}  
$S_{b}$ is the per-block score from the LLM evaluator; $\bar{S}$ is the mean score across blocks.

\textbf{Computation:}  
Implemented via asynchronous calls to the \texttt{gpt-4.1-mini} model through the \texttt{g4f} library. A structured prompt template was used to present the original source text alongside the extracted texts for each block. The prompt was designed to be analogous to the instructions provided to human experts. The LLM was instructed to score each block on a $1$–$5$ scale based on fidelity to the source, providing a brief reason for each score. Responses were parsed as JSON; invalid responses triggered retries (up to three attempts) with error feedback included in the prompt. Scores and reasons were stored in the metrics dictionary for each block.

\textbf{Relevance:}  
Assesses the task-specific accuracy and completeness of extractions, ensuring semantic and contextual alignment with the original document.

\textbf{Strengths:}  
Provides granular, interpretable feedback with reasons; mimics human-like qualitative assessment.

\textbf{Limitations:}  
Subjective to the evaluating LLM's judgment; potential variability across model runs or providers.

\subsubsection{Structural Metrics}
\paragraph{Block Order Consistency}
\textbf{Purpose:} Measures the alignment of token order in $E$ with occurrences in $D$.
\textbf{Formulation:}
$$\text{Block Order Consistency} = \frac{\tau + 1}{2}
\label{eq12}$$
\textbf{Notation:} $\tau$ is Kendall’s Tau correlation between the positions of matching tokens in the sequence of all blocks ($E$) and their first unused occurrences in $D$. Positions in $E$ are sequential indices in the tokenized $E$. If fewer than 2 matching tokens, score is 0.
\textbf{Computation:} Use scipy.stats.kendalltau and normalize: $(\tau + 1)/2$. Tokens are lemmatized and matched greedily, using the next available index in $D$ greater than previous matches.
\textbf{Relevance:} Ensures that the judicial narrative flow is preserved.
\textbf{Strengths:} Formalized, interpretable, automatically computable.
\textbf{Limitations:} Greedy matching may not be optimal; ignores duplicates perfectly.
\paragraph{Block Completeness}
\textbf{Purpose:} Assesses key term coverage in $E$.
\textbf{Formulation:}
$$\text{Block Completeness} = \frac{|K_D \cap K_E|}{|K_D|}
\label{eq13}$$
\textbf{Notation:} $K_D$ is the set of the top-50 TF-IDF terms from the original document $D$. $K_E$ is the set of lemmatized tokens in the combined blocks $E$. $K_D \cap K_E$ is the intersection of terms. $|K_D|$ is the number of terms in $K_D$.
\textbf{Computation:} Use TF-IDF terms.
\textbf{Relevance:} Ensures comprehensive coverage.
\textbf{Strengths:} Granular.
\textbf{Limitations:} Term-based; identical to Coverage Ratio in current implementation.
\paragraph{Monotonicity Score}
\textbf{Purpose:} Ensures chronological date order in $E$.
\textbf{Formulation:}
$$\text{Monotonicity Score} = \begin{cases}
1 & \text{if } T_E = \text{sort}(T_E) \\
0 & \text{otherwise}
\end{cases}
\label{eq14}$$
\textbf{Notation:} $T_E$ is the list of dates extracted from the combined blocks $E$ in order of appearance, formatted as YYYY-MM-DD. $\text{sort}(T_E)$ is the chronologically sorted order of these dates. The score is 1 if the extracted dates appear in chronological order, otherwise 0.
\textbf{Computation:} Use Natasha's DatesExtractor.
\textbf{Relevance:} Preserves case timelines.
\textbf{Strengths:} Domain-specific.
\textbf{Limitations:} Only checks order in $E$, not coverage from $D$; misses implicit references.
\subsubsection{Pseudo-Ground Truth Metrics}
\paragraph{Keyword-Based Pseudo-F1}
\textbf{Purpose:} Mimics F1 using original key terms.
\textbf{Formulation:}
\begin{align}
P &= \frac{|K_D \cap K_E|}{|K_E|}, \quad R = \frac{|K_D \cap K_E|}{|K_D|}, \quad F1 = \frac{2PR}{P + R}
\label{eq15}
\end{align}
\textbf{Notation:} $K_D$ is the set of the top-50 TF-IDF terms from the original document $D$. $K_E$ is the set of lemmatized tokens from the combined extracted blocks $E$. $K_D \cap K_E$ is the intersection of terms. $P$ (precision) is the proportion of extracted terms that are key terms. $R$ (recall) is the proportion of key terms captured in $E$. $F1$ is the harmonic mean of precision and recall.
\textbf{Computation:} Use TF-IDF terms.
\textbf{Relevance:} Supervised-like evaluation.
\textbf{Strengths:} Robust proxy.
\textbf{Limitations:} Term-based.
\subsubsection{Legal-Specific Metrics}
\paragraph{Legal Term Density}  
\textbf{Purpose:} Measures the concentration of legal terms within a single block.  
\textbf{Formulation:}  
\[
\text{Legal Term Density}_{B_i} = \frac{|C_E \cap K_{B_i}|}{|K_{B_i}|}, \quad i=1\dots 7
\]  
\textbf{Notation:} $C_E$ is the set of legal entities (ORG, PER, LOC via NER), legal patterns, and top-20 TF-IDF terms from the combined document. $K_{B_i}$ is the set of lemmatized tokens in block $B_i$. The ratio measures the concentration of legal terms specifically in block $B_i$.  
\textbf{Computation:} Calculated per block using tokenization and custom citation/legal term extraction~\cite{araujo2019citations}; normalized as needed.  
\textbf{Relevance:} Highlights the legal focus of each block.  
\textbf{Strengths:} Simple and interpretable.  
\textbf{Limitations:} Term-based; may overlap with other metrics like Coverage Ratio or Block Completeness.

\section{Correlation and Agreement Computation Analysis between Expert Ratings and Unsupervised Metrics}\label{sec:correlations}

To evaluate the alignment of unsupervised metrics with human judgment, we computed correlations between metric scores and expert ratings using a robust statistical pipeline implemented in Python 3.12. The analysis is described below.

\subsection{Data Preparation}

We used the \texttt{merged_output_with_expert_eval_all.json} file, which contains 1,000 judicial decisions with associated expert ratings and computed metric scores. Each document consists of seven semantic blocks (plaintiff claims, plaintiff arguments, defendant arguments, court evidence evaluation, reasoning steps, applicable norms, and final ruling), with both block-level and document-level metrics. Expert ratings are provided on a 1--5 Likert scale.

\subsection{Aggregation}

\begin{itemize}
\item \textbf{Block-level metrics}: For each block, metrics were collected and paired with the corresponding expert evaluations for the same block. When multiple experts rated the same block, their scores were averaged to compute the block-level expert rating.
\item \textbf{Document-level metrics}: Metrics computed over the entire document were paired with the \emph{average of all block-level expert ratings} for that document.
\end{itemize}

\subsection{Normalization}

All metric values and expert ratings were normalized using min--max scaling, computed globally per metric to ensure comparability across blocks and documents. For constant metrics (min = max), a default normalized value of 0.5 was assigned.

\subsection{Correlation and Agreement Computation}

We computed correlation and agreement measures between each metric and its corresponding expert ratings:

\begin{itemize}
\item \textbf{Pearson correlation} (\texttt{pearson r}): measures linear relationships.
\item \textbf{Spearman correlation} (\texttt{spearman r}): evaluates monotonic relationships.
\item \textbf{Kendall's Tau} (\texttt{kendalltau}): quantifies ordinal association.
\item \textbf{Lin's Concordance Correlation Coefficient (CCC)}: quantifies the agreement between normalized metric and expert scores, accounting for both precision and accuracy.
\item \textbf{Mean Absolute Error (MAE)}: measures the average absolute difference between normalized metric and expert scores.
\end{itemize}

For block-level metrics, correlations and agreement measures were computed per block using the block-specific expert ratings. For document-level metrics, they were computed against the averaged document-level expert ratings. All computations used only valid (non-missing) values, and metrics were optionally bootstrapped to estimate 95

\subsection{Inter-Expert Agreement}

To assess consistency among human raters, we computed the intraclass correlation coefficient ICC(2,k)~\cite{shrout1979} for blocks or documents with the modal number of raters. Normalized expert scores were used, and a one-way or two-way ANOVA model was fitted to obtain the ICC values. This provides a measure of reliability of human evaluations across multiple raters.

\subsection{Results Storage and Post-Processing}

All correlation coefficients, summary statistics, ICC estimates, and expert statistics were serialized to a JSON file (\texttt{metrics_analysis.json}) for reproducibility and downstream analysis. Extreme or missing values were filtered using checks and regular expressions.

\section{Results and discussion }\label{sec:results}

We evaluated 16 unsupervised metrics on the \texttt{sud-resh-benchmark} dataset, which comprises 1,000 anonymized Russian judicial decisions stratified across ten legal areas. These decisions were annotated with 7168 expert reviews on a 1--5 Likert scale. The metrics target the extraction of seven semantic blocks: plaintiff's claims, plaintiff's arguments, defendant's arguments, court evaluation of evidence, judge's reasoning steps and intermediate conclusions, applicable legal norms, and the final court decision.

We analyzed both block-level metrics (\emph{Intra-Block Coherence}, \emph{Inter-Block Distinctiveness}, \emph{Neutrality Bias}, \emph{Legal Term Density}, \emph{LLM Evaluation Score}) and document-level metrics (\emph{Coverage Ratio}, \emph{Redundancy Penalty}, \emph{Compression Ratio}, \emph{Term Frequency Coherence}, \emph{Citation Coverage}, \emph{Semantic Entropy}, \emph{Raw Cosine Similarity}, \emph{Block Order Consistency}, \emph{Monotonicity Score}, \emph{Block Completeness}, \emph{Keyword-Based Pseudo-F1}) using descriptive statistics (mean, variance, standard deviation) and correlations (Pearson, Spearman, Kendall) with averaged expert ratings, as well as Lin's concordance correlation coefficient (CCC) and mean absolute error (MAE).

To highlight the key results, we provide a concise table. Table~\ref{tab:metrics_summary} summarizes the block-level and document-level metrics, presenting their mean, Pearson correlation, Lin's CCC, and MAE to reflect alignment with expert judgments.

\begin{table}[h!]
\centering
\caption{Summary of Block-Level and Document-Level Metrics Statistics}
\label{tab:metrics_summary}
\small
\begin{tabular}{|m{4cm}|c|c|c|c|}
\hline
\textbf{Metric} & \textbf{Mean} & \textbf{Pearson r} & \textbf{Lin CCC} & \textbf{MAE} \\
\hline
Intra-Block Coherence & 0.895 & -0.207 & -0.142 & 0.196 \\
Inter-Block Distinctiveness & 0.366 & -0.404 & -0.065 & 0.421 \\
Neutrality Bias & 0.500 & -- & -- & 0.286 \\
Legal Term Density & 0.398 & -0.479 & -0.079 & 0.394 \\
LLM Evaluation Score & 0.849 & 0.382 & 0.325 & 0.197 \\
Coverage Ratio & 0.685 & 0.513 & 0.443 & 0.139 \\
Redundancy Penalty & 0.512 & -0.404 & -0.175 & 0.309 \\
Compression Ratio & 0.274 & 0.412 & 0.069 & 0.497 \\
Term Freq. Coherence & 0.720 & 0.540 & 0.512 & 0.127 \\
Citation Coverage & 0.556 & 0.378 & 0.194 & 0.232 \\
Semantic Entropy & 0.711 & 0.444 & 0.381 & 0.119 \\
Raw Cosine Similarity & 0.603 & 0.207 & 0.115 & 0.203 \\
Block Order Consistency & 0.680 & 0.245 & 0.191 & 0.149 \\
Monotonicity Score & 0.500 & -- & -- & 0.286 \\
Block Completeness & 0.685 & 0.513 & 0.443 & 0.139 \\
Keyword-Based Pseudo-F1 & 0.634 & 0.308 & 0.203 & 0.173 \\
\hline
\end{tabular}
\end{table}
\subsection*{Block-Level Metrics}

At the block level ($n = 6{,}965$ evaluations), \emph{Intra-Block Coherence} had a high mean of $0.895$ (variance $0.033$, std. dev. $0.182$) but showed a negative correlation with expert ratings (Pearson $r = -0.207$, $p = 2.59 \times 10^{-6}$; Spearman $r = -0.223$, $p = 3.69 \times 10^{-7}$; Kendall $r = -0.162$, $p = 4.26 \times 10^{-7}$; Lin CCC $ = -0.142$; MAE $0.196$). This suggests that overly cohesive blocks may not align with expert preferences.

\emph{Inter-Block Distinctiveness} (mean $= 0.366$, variance $0.020$, std. dev. $0.142$) exhibited an even stronger negative correlation (Pearson $r = -0.404$, $p = 2.34 \times 10^{-21}$; Spearman $r = -0.389$, $p = 8.02 \times 10^{-20}$; Kendall $r = -0.272$, $p = 3.45 \times 10^{-19}$; Lin CCC $ = -0.065$; MAE $0.421$).

\emph{Neutrality Bias} remained constant at $1$ ($0.5$ in Appendix~\ref{secA1}) with zero variance, reflecting no detectable sentiment bias but also highlighting potential limitations of sentiment analysis models for legal texts.

\emph{Legal Term Density} (mean $= 0.398$, variance $0.024$, std. dev. $0.156$) demonstrated the strongest negative correlation (Pearson $r = -0.479$, $p = 1.38 \times 10^{-30}$; Spearman $r = -0.353$, $p = 2.12 \times 10^{-16}$; Kendall $r = -0.249$, $p = 2.77 \times 10^{-16}$; Lin CCC $ = -0.079$; MAE $0.394$), implying that higher concentrations of legal terminology might compromise clarity or relevance from the experts’ perspective.

The \emph{LLM Evaluation Score} (mean $= 0.849$, variance $0.055$, std. dev. $0.235$) showed a positive but modest correlation (Pearson $r = 0.382$, $p = 4.07 \times 10^{-19}$; Spearman $r = 0.247$, $p = 1.59 \times 10^{-8}$; Kendall $r = 0.190$, $p = 2.31 \times 10^{-8}$; Lin CCC $ = 0.325$; MAE $0.197$). This suggests that while large language models capture certain aspects of extract quality, the model used may not be sufficiently specialized for precise legal assessment.

\subsection*{Document-Level Metrics}

At the document level ($n = 995$ evaluations), \emph{Term Frequency Coherence} (mean $= 0.720$, variance $0.028$, std. dev. $0.168$) achieved the highest positive correlation with expert ratings (Pearson $r = 0.540$, $p = 8.30 \times 10^{-40}$; Spearman $r = 0.320$, $p = 1.33 \times 10^{-13}$; Kendall $r = 0.222$, $p = 3.16 \times 10^{-13}$; Lin CCC $ = 0.512$; MAE $0.127$), indicating strong alignment between term distribution in extracts and expert judgments.

\emph{Coverage Ratio} and \emph{Block Completeness} (both mean $= 0.685$, variance $0.026$, std. dev. $0.163$) followed closely (Pearson $r = 0.513$, $p = 1.50 \times 10^{-35}$; Spearman $r = 0.353$, $p = 2.18 \times 10^{-16}$; Kendall $r = 0.253$, $p = 3.17 \times 10^{-16}$; Lin CCC $ = 0.443$; MAE $0.139$), emphasizing the importance of capturing key terms.

\emph{Semantic Entropy} (mean $= 0.711$, variance $0.011$, std. dev. $0.106$) and \emph{Compression Ratio} (mean $= 0.274$, variance $0.030$, std. dev. $0.173$) both showed positive correlations with expert ratings. The associations were substantially stronger for \emph{Semantic Entropy} (Pearson $r = 0.444$, $p = 4.79 \times 10^{-26}$; Spearman $r = 0.364$, $p = 1.95 \times 10^{-17}$; Kendall $r = 0.255$, $p = 4.84 \times 10^{-17}$; Lin CCC $= 0.381$; MAE $= 0.119$), highlighting its informativeness. In contrast, \emph{Compression Ratio} exhibited weaker relationships (Pearson $r = 0.412$, $p = 2.89 \times 10^{-22}$; Spearman $r = 0.390$, $p = 6.48 \times 10^{-20}$; Kendall $r = 0.276$, $p = 1.09 \times 10^{-19}$; Lin CCC $= 0.069$; MAE $= 0.497$), suggesting its limited ability to capture extraction quality. Overall, \emph{Semantic Entropy} emerged as the more reliable metric, whereas \emph{Compression Ratio} demonstrated only a weak alignment with expert judgments.

In contrast, \emph{Redundancy Penalty} (mean $= 0.512$, variance $0.030$, std. dev. $0.173$) was negatively correlated (Pearson $r = -0.404$, $p = 2.34 \times 10^{-21}$; Spearman $r = -0.389$, $p = 8.02 \times 10^{-20}$; Kendall $r = -0.272$, $p = 3.45 \times 10^{-19}$; Lin CCC $ = -0.175$; MAE $0.309$), reflecting a penalty for overlapping content.

\emph{Raw Cosine Similarity} (mean $= 0.603$, variance $0.015$, std. dev. $0.123$) correlated poorly with expert ratings (Pearson $r = 0.207$, $p = 2.43 \times 10^{-6}$; Spearman $r = 0.082$, $p = 0.06$; Kendall $r = 0.055$, $p = 0.07$; Lin CCC $ = 0.115$; MAE $0.203$), likely due to the peculiarities of the Russian language and the use of a general-purpose vectorization model rather than one fine-tuned for legal texts.

Other metrics like \emph{Citation Coverage} (mean $= 0.556$, variance $0.026$, std. dev. $0.162$; Pearson $r = 0.378$, $p = 9.57 \times 10^{-19}$; Lin CCC $ = 0.194$; MAE $0.232$), \emph{Block Order Consistency} (mean $= 0.680$, variance $0.013$, std. dev. $0.115$; Pearson $r = 0.245$, $p = 2.30 \times 10^{-8}$; Lin CCC $ = 0.191$; MAE $0.149$), and \emph{Keyword-Based Pseudo-F1} (mean $= 0.634$, variance $0.016$, std. dev. $0.125$; Pearson $r = 0.308$, $p = 1.29 \times 10^{-12}$; Lin CCC $ = 0.203$; MAE $0.173$) showed varying degrees of alignment, with generally moderate correlations and CCC values underscoring the need for domain-specific refinements.

\subsection*{Per-Block Variations}

Examining per-block variations revealed additional insights. For \emph{Intra-Block Coherence}, the \emph{Court Evidence Evaluation} block exhibited the highest mean ($0.930$, variance $0.020$, std. dev. $0.140$) but a strong negative correlation with expert ratings (Pearson $r = -0.340$, $p = 3.53 \times 10^{-15}$; Lin CCC $ = -0.140$; MAE $0.376$). By contrast, \emph{Legal Norms} displayed the highest mean for \emph{Inter-Block Distinctiveness} ($0.444$, variance $0.025$, std. dev. $0.157$; Pearson $r = -0.194$, $p = 1.09 \times 10^{-5}$; Lin CCC $ = -0.110$; MAE $0.307$).

For \emph{Legal Term Density}, means ranged from $0.376$ (\emph{Defendant's Arguments}) to $0.434$ (\emph{Court Decision}), with negative correlations strongest in \emph{Judge's Reasoning} (Pearson $r = -0.367$, $p = 1.33 \times 10^{-17}$; Lin CCC $ = -0.188$; MAE $0.345$).

\emph{LLM Evaluation Score} showed means from $0.829$ (\emph{Legal Norms}) to $0.871$ (\emph{Court Decision}), with positive correlations across blocks, strongest in \emph{Plaintiff's Arguments} and \emph{Court Decision} (Pearson $r \approx 0.283-0.286$; Lin CCC up to $0.282$; MAE as low as $0.142$).

Overall, correlations were moderate; for instance, $r = 0.54$ for \emph{Term Frequency Coherence} may be considered relatively low for high-stakes legal applications, where even minor errors can have significant consequences. Lin CCC values, often below $0.5$, and MAE metrics further indicate room for improvement in metric agreement with experts. These findings underscore that unsupervised metrics, while useful for initial screening or scalability, cannot fully substitute for human judgment. Expert evaluations remain essential for accurate assessment of information extraction in legal contexts.

\subsection*{Comparative Performance of Semantic Metrics}

Among the semantic metrics, \emph{Raw Cosine Similarity} exhibited a modest mean value of $0.603$ (variance $0.015$, std. dev. $0.123$) but demonstrated only a weak positive correlation with expert ratings (Pearson $r = 0.207$, $p = 2.43 \times 10^{-6}$; Lin CCC $ = 0.115$; MAE $0.203$), underscoring its limited validity as a standalone proxy for extraction quality. This underperformance likely stems from the use of a general-purpose SentenceTransformer model (\texttt{all-MiniLM-L6-v2}), which is not fine-tuned for Russian legal texts, leading to suboptimal capture of domain-specific nuances such as jurisdictional terminology, formal phrasing, and contextual interdependencies prioritized by experts.

Similarly, the \emph{LLM Evaluation Score} achieved a higher mean of $0.849$ (variance $0.055$, std. dev. $0.235$) and a moderate positive correlation (Pearson $r = 0.382$, $p = 4.07 \times 10^{-19}$; Lin CCC $ = 0.325$; MAE $0.197$). Nevertheless, its modest alignment, as evidenced by the CCC below $0.4$ and non-negligible MAE, suggests that the evaluating model (\texttt{gpt-4.1-mini} via \texttt{g4f}) lacks sufficient specialization for legal assessment, potentially introducing subjectivity or overlooking subtle fidelity issues such as argumentative coherence.

To enhance these metrics, future iterations should incorporate domain-adapted embeddings (e.g., Legal-BERT variants trained on Russian jurisprudence) or fine-tuned LLMs prompted with legal-specific rubrics. Such adaptations would likely boost semantic fidelity, correlation strengths, and CCC while reducing MAE.

While these unsupervised approaches demand cautious interpretation and rigorous preprocessing—particularly for multilingual or jargon-heavy corpora—they offer excellent scalability for processing vast judicial archives without exhaustive annotations. A promising hybrid paradigm emerges from leveraging expert annotations to curate training data for fine-tuning encoder models or LLMs: initial human oversight ensures contextual accuracy, enabling subsequent autonomous handling of large-scale datasets and fostering ethical, high-fidelity AI in legal NLP.

Detailed statistics are provided in Appendix~\ref{secA1}.

\section{Implementation}\label{sec7}
Metrics are implemented in Python 3.12 with dependencies listed above. Below is the Coverage Ratio implementation:
\begin{lstlisting}
import spacy
from sklearn.feature_extraction.text import TfidfVectorizer
from nltk.corpus import stopwords
nlp = spacy.load("ru_core_news_sm")
stop_words = set(stopwords.words("russian"))
def tokenize(text):
return [token.lemma_ for token in nlp(text.lower()) if token.text not in stop_words and token.is_alpha]
def custom_analyzer(text):
return tokenize(text)
def get_key_terms(text, n=50):
vectorizer = TfidfVectorizer(analyzer=custom_analyzer)
tfidf = vectorizer.fit_transform([text])
terms = vectorizer.get_feature_names_out()
scores = tfidf.toarray()[0]
return [terms[i] for i in scores.argsort()[-n:]]
def coverage_ratio(original, extracted):
key_terms = get_key_terms(original, n=50)
extracted_terms = set(tokenize(extracted))
return len(set(key_terms) & extracted_terms) / len(key_terms) if key_terms else 0
\end{lstlisting}
Code and expert data are available at \url{https://github.com/TryDotAtwo/TestEvalForLaw}.
\section{Conclusion}\label{sec12}

This study evaluated 16 unsupervised metrics for extracting seven semantic blocks from 1,000 anonymized Russian judicial decisions, validated against 7,168 expert reviews on a 1--5 Likert scale. Among document-level metrics, \emph{Term Frequency Coherence} (mean = 0.720, Lin CCC = 0.512, MAE = 0.127) and \emph{Coverage Ratio}/\emph{Block Completeness} (mean = 0.685, Lin CCC = 0.443, MAE = 0.139) demonstrated the strongest alignment with expert ratings (Pearson $r = 0.540$, $p = 8.30 \times 10^{-40}$ and $r = 0.513$, $p = 1.50 \times 10^{-35}$, respectively). \emph{Semantic Entropy} (mean = 0.711, Lin CCC = 0.381, MAE = 0.119) also showed robust performance, underscoring the value of diverse content in extracts. Conversely, \emph{Legal Term Density} (mean = 0.398, Lin CCC = -0.079, MAE = 0.394) and \emph{Inter-Block Distinctiveness} (mean = 0.366, Lin CCC = -0.065, MAE = 0.421) exhibited strong negative correlations (Pearson $r = -0.479$, $p = 1.38 \times 10^{-30}$ and $r = -0.404$, $p = 2.34 \times 10^{-21}$), suggesting that excessive legal jargon or block separation may reduce clarity or relevance. \emph{Neutrality Bias} (mean = 0.500, MAE = 0.286) and \emph{Monotonicity Score} (mean = 0.500, MAE = 0.286) were constant and thus ineffective due to zero variance.

The \emph{LLM Evaluation Score} (mean = 0.849, Lin CCC = 0.325, MAE = 0.197) showed a moderate positive correlation (Pearson $r = 0.382$, $p = 4.07 \times 10^{-19}$), but its modest Lin CCC and non-negligible MAE indicate that the evaluating model (\texttt{gpt-4.1-mini} via \texttt{g4f}) lacks sufficient specialization for legal texts, potentially missing nuances like argumentative coherence. Similarly, \emph{Raw Cosine Similarity} (mean = 0.603, Lin CCC = 0.115, MAE = 0.203) performed poorly (Pearson $r = 0.207$, $p = 2.43 \times 10^{-6}$), likely due to the general-purpose SentenceTransformer model (\texttt{all-MiniLM-L6-v2}) not being fine-tuned for Russian legal texts.

These findings highlight that while unsupervised metrics offer scalability for processing large judicial archives, their moderate correlations (e.g., highest Pearson $r = 0.540$) and generally low Lin CCC values (often < 0.5) underscore the need for human oversight in high-stakes legal applications where precision is critical. Future improvements should leverage domain-adapted embeddings (e.g., Legal-BERT trained on Russian jurisprudence) or fine-tuned LLMs with legal-specific rubrics to enhance semantic fidelity, correlation strength, and agreement with expert judgments while reducing MAE. A hybrid approach—combining expert annotations for initial model training with automated processing—promises to balance accuracy and scalability, fostering ethical, high-fidelity AI applications in legal NLP.

Detailed statistics are provided in Appendix~\ref{secA1}.

\backmatter

\section*{Declarations}
\begin{itemize}     \item Funding: Not applicable     \item Conflict of interest/Competing interests: The authors declare no competing interests.     \item Ethics approval and consent to participate: Not applicable     \item Consent for publication: Not applicable     \item Data availability: The data that support the findings of this study are openly available at \url{https://huggingface.co/datasets/lawful-good-project/sud-resh-benchmark} and in \url{https://github.com/TryDotAtwo/TestEvalForLaw}.     \item Materials availability: Not applicable     \item Code availability: The code is available at \url{https://github.com/TryDotAtwo/TestEvalForLaw}.     \item Author contribution: Ivan Litvak and Anton Kostin conceptualized the study, designed the unsupervised metrics, implemented the code, conducted the experiments, and analyzed the results. Sergey Lagutin, Tatiana Maksiyan, and Fedor Lashkin served as legal experts, providing detailed evaluations for 7 168 reviews of judicial decisions, validating the metric performance against human judgments, and contributing to the interpretation of findings in the legal context. All authors contributed to the writing and revision of the manuscript and approved the final version.     \end{itemize}


\bibliography{sn-bibliography}

\begin{appendices}

\section{Detailed Statistical Tables}\label{secA1}

\begin{table}[htbp]
\centering
\caption{Global Statistics for Block-Level Metrics (Part 1)}
\label{tab:block_global_part1}
\small
\begin{tabular}{|m{2.2cm}|c|c|c|c|}
\hline
\textbf{Metric} & \textbf{\# Eval.} & \textbf{Mean} & \textbf{Var.} & \textbf{Std. Dev.} \\
\hline
Intra-Block Coherence & 6965 & 0.895 & 0.033 & 0.182 \\
Inter-Block Distinctiveness & 6965 & 0.366 & 0.020 & 0.142 \\
Neutrality Bias & 6965 & 0.500 & 0.000 & 0.000 \\
Legal Term Density & 6965 & 0.398 & 0.024 & 0.156 \\
LLM Evaluation Score & 6965 & 0.849 & 0.055 & 0.235 \\
\hline
\end{tabular}
\end{table}

\begin{table}[htbp]
\centering
\caption{Global Statistics for Block-Level Metrics (Part 2)}
\label{tab:block_global_part2}
\small
\begin{tabular}{|m{2.2cm}|m{1.8cm}|m{1.8cm}|m{1.8cm}|c|c|}
\hline
\textbf{Metric} & \textbf{Pearson r (p)} & \textbf{Spearman r (p)} & \textbf{Kendall r (p)} & \textbf{Lin CCC} & \textbf{MAE} \\
\hline
Intra-Block Coherence & -0.207 ($2.59 \times 10^{-6}$) & -0.223 ($3.69 \times 10^{-7}$) & -0.162 ($4.26 \times 10^{-7}$) & -0.142 & 0.196 \\
Inter-Block Distinctiveness & -0.404 ($2.34 \times 10^{-21}$) & -0.389 ($8.02 \times 10^{-20}$) & -0.272 ($3.45 \times 10^{-19}$) & -0.065 & 0.421 \\
Neutrality Bias & -- & -- & -- & -- & 0.286 \\
Legal Term Density & -0.479 ($1.38 \times 10^{-30}$) & -0.353 ($2.12 \times 10^{-16}$) & -0.249 ($2.77 \times 10^{-16}$) & -0.079 & 0.394 \\
LLM Evaluation Score & 0.382 ($4.07 \times 10^{-19}$) & 0.247 ($1.59 \times 10^{-8}$) & 0.190 ($2.31 \times 10^{-8}$) & 0.325 & 0.197 \\
\hline
\end{tabular}
\end{table}

\begin{table}[htbp]
\centering
\caption{Per-Block Statistics for Intra-Block Coherence (Part 1)}
\label{tab:intra-block_coherence_per_block_part1}
\small
\begin{tabular}{|m{2.2cm}|c|c|c|c|c|}
\hline
\textbf{Block} & \textbf{\# Eval.} & \textbf{Mean} & \textbf{Var.} & \textbf{Std. Dev.} & \textbf{Avg. Experts} \\
\hline
Plaintiff's Claims & 995 & 0.897 & 0.035 & 0.187 & 2.028 \\
Plaintiff's Arguments & 995 & 0.862 & 0.037 & 0.192 & 2.024 \\
Defendant's Arguments & 995 & 0.915 & 0.027 & 0.163 & 2.018 \\
Court Evidence Eval. & 995 & 0.930 & 0.020 & 0.140 & 2.010 \\
Judge's Reasoning & 995 & 0.896 & 0.028 & 0.169 & 2.008 \\
Legal Norms & 995 & 0.891 & 0.046 & 0.213 & 2.014 \\
Court Decision & 995 & 0.873 & 0.036 & 0.190 & 2.018 \\
\hline
\end{tabular}
\end{table}

\begin{table}[htbp]
\centering
\caption{Per-Block Statistics for Intra-Block Coherence (Part 2)}
\label{tab:intra-block_coherence_per_block_part2}
\small
\begin{tabular}{|m{2.2cm}|m{1.8cm}|m{1.8cm}|m{1.8cm}|c|c|}
\hline
\textbf{Block} & \textbf{Pearson r (p)} & \textbf{Spearman r (p)} & \textbf{Kendall r (p)} & \textbf{Lin CCC} & \textbf{MAE} \\
\hline
Plaintiff's Claims & -0.045 (0.31) & -0.052 (0.24) & -0.045 (0.25) & -0.044 & 0.181 \\
Plaintiff's Arguments & -0.024 (0.59) & -0.032 (0.47) & -0.027 (0.46) & -0.024 & 0.212 \\
Defendant's Arguments & -0.083 (0.06) & -0.129 ($3.71 \times 10^{-3}$) & -0.108 ($3.96 \times 10^{-3}$) & -0.066 & 0.244 \\
Court Evidence Eval. & -0.340 ($3.53 \times 10^{-15}$) & -0.367 ($1.38 \times 10^{-17}$) & -0.307 ($1.58 \times 10^{-16}$) & -0.140 & 0.376 \\
Judge's Reasoning & -0.249 ($1.30 \times 10^{-8}$) & -0.292 ($1.90 \times 10^{-11}$) & -0.242 ($3.02 \times 10^{-11}$) & -0.150 & 0.329 \\
Legal Norms & -0.163 ($2.23 \times 10^{-4}$) & -0.175 ($7.05 \times 10^{-5}$) & -0.146 ($8.59 \times 10^{-5}$) & -0.106 & 0.348 \\
Court Decision & -0.009 (0.84) & -0.024 (0.59) & -0.021 (0.59) & -0.008 & 0.154 \\
\hline
\end{tabular}
\end{table}

\begin{table}[htbp]
\centering
\caption{Per-Block Statistics for Inter-Block Distinctiveness (Part 1)}
\label{tab:inter-block_distinctiveness_per_block_part1}
\small
\begin{tabular}{|m{2.2cm}|c|c|c|c|c|}
\hline
\textbf{Block} & \textbf{\# Eval.} & \textbf{Mean} & \textbf{Var.} & \textbf{Std. Dev.} & \textbf{Avg. Experts} \\
\hline
Plaintiff's Claims & 995 & 0.351 & 0.018 & 0.135 & 2.028 \\
Plaintiff's Arguments & 995 & 0.354 & 0.017 & 0.128 & 2.024 \\
Defendant's Arguments & 995 & 0.345 & 0.019 & 0.138 & 2.018 \\
Court Evidence Eval. & 995 & 0.394 & 0.023 & 0.151 & 2.010 \\
Judge's Reasoning & 995 & 0.337 & 0.015 & 0.122 & 2.008 \\
Legal Norms & 995 & 0.444 & 0.025 & 0.157 & 2.014 \\
Court Decision & 995 & 0.334 & 0.016 & 0.126 & 2.018 \\
\hline
\end{tabular}
\end{table}

\begin{table}[htbp]
\centering
\caption{Per-Block Statistics for Inter-Block Distinctiveness (Part 2)}
\label{tab:inter-block_distinctiveness_per_block_part2}
\small
\begin{tabular}{|m{2.2cm}|m{1.8cm}|m{1.8cm}|m{1.8cm}|c|c|}
\hline
\textbf{Block} & \textbf{Pearson r (p)} & \textbf{Spearman r (p)} & \textbf{Kendall r (p)} & \textbf{Lin CCC} & \textbf{MAE} \\
\hline
Plaintiff's Claims & -0.262 ($2.20 \times 10^{-9}$) & -0.190 ($1.68 \times 10^{-5}$) & -0.145 ($1.42 \times 10^{-5}$) & -0.044 & 0.546 \\
Plaintiff's Arguments & -0.187 ($2.19 \times 10^{-5}$) & -0.159 ($3.09 \times 10^{-4}$) & -0.119 ($3.14 \times 10^{-4}$) & -0.035 & 0.500 \\
Defendant's Arguments & -0.290 ($2.68 \times 10^{-11}$) & -0.286 ($5.34 \times 10^{-11}$) & -0.212 ($6.50 \times 10^{-11}$) & -0.073 & 0.468 \\
Court Evidence Eval. & -0.476 ($5.73 \times 10^{-30}$) & -0.459 ($8.55 \times 10^{-28}$) & -0.343 ($8.94 \times 10^{-27}$) & -0.248 & 0.348 \\
Judge's Reasoning & -0.381 ($5.46 \times 10^{-19}$) & -0.382 ($4.40 \times 10^{-19}$) & -0.283 ($1.74 \times 10^{-18}$) & -0.123 & 0.387 \\
Legal Norms & -0.194 ($1.09 \times 10^{-5}$) & -0.190 ($1.62 \times 10^{-5}$) & -0.139 ($1.81 \times 10^{-5}$) & -0.110 & 0.307 \\
Court Decision & -0.136 ($2.13 \times 10^{-3}$) & -0.114 (0.01) & -0.090 (0.01) & -0.013 & 0.627 \\
\hline
\end{tabular}
\end{table}

\begin{table}[htbp]
\centering
\caption{Per-Block Statistics for Neutrality Bias (Part 1)}
\label{tab:neutrality_bias_per_block_part1}
\small
\begin{tabular}{|m{2.2cm}|c|c|c|c|c|}
\hline
\textbf{Block} & \textbf{\# Eval.} & \textbf{Mean} & \textbf{Var.} & \textbf{Std. Dev.} & \textbf{Avg. Experts} \\
\hline
Plaintiff's Claims & 995 & 0.500 & 0.000 & 0.000 & 2.028 \\
Plaintiff's Arguments & 995 & 0.500 & 0.000 & 0.000 & 2.024 \\
Defendant's Arguments & 995 & 0.500 & 0.000 & 0.000 & 2.018 \\
Court Evidence Eval. & 995 & 0.500 & 0.000 & 0.000 & 2.010 \\
Judge's Reasoning & 995 & 0.500 & 0.000 & 0.000 & 2.008 \\
Legal Norms & 995 & 0.500 & 0.000 & 0.000 & 2.014 \\
Court Decision & 995 & 0.500 & 0.000 & 0.000 & 2.018 \\
\hline
\end{tabular}
\end{table}

\begin{table}[htbp]
\centering
\caption{Per-Block Statistics for Neutrality Bias (Part 2)}
\label{tab:neutrality_bias_per_block_part2}
\small
\begin{tabular}{|m{2.2cm}|m{1.8cm}|m{1.8cm}|m{1.8cm}|c|c|}
\hline
\textbf{Block} & \textbf{Pearson r (p)} & \textbf{Spearman r (p)} & \textbf{Kendall r (p)} & \textbf{Lin CCC} & \textbf{MAE} \\
\hline
Plaintiff's Claims & -- & -- & -- & -- & 0.397 \\
Plaintiff's Arguments & -- & -- & -- & -- & 0.354 \\
Defendant's Arguments & -- & -- & -- & -- & 0.318 \\
Court Evidence Eval. & -- & -- & -- & -- & 0.225 \\
Judge's Reasoning & -- & -- & -- & -- & 0.237 \\
Legal Norms & -- & -- & -- & -- & 0.213 \\
Court Decision & -- & -- & -- & -- & 0.464 \\
\hline
\end{tabular}
\end{table}

\begin{table}[htbp]
\centering
\caption{Per-Block Statistics for Legal Term Density (Part 1)}
\label{tab:legal_term_density_per_block_part1}
\small
\begin{tabular}{|m{2.2cm}|c|c|c|c|c|}
\hline
\textbf{Block} & \textbf{\# Eval.} & \textbf{Mean} & \textbf{Var.} & \textbf{Std. Dev.} & \textbf{Avg. Experts} \\
\hline
Plaintiff's Claims & 995 & 0.416 & 0.027 & 0.164 & 2.028 \\
Plaintiff's Arguments & 995 & 0.377 & 0.021 & 0.144 & 2.024 \\
Defendant's Arguments & 995 & 0.376 & 0.022 & 0.147 & 2.018 \\
Court Evidence Eval. & 995 & 0.377 & 0.027 & 0.164 & 2.010 \\
Judge's Reasoning & 995 & 0.421 & 0.024 & 0.154 & 2.008 \\
Legal Norms & 995 & 0.386 & 0.024 & 0.153 & 2.014 \\
Court Decision & 995 & 0.434 & 0.024 & 0.155 & 2.018 \\
\hline
\end{tabular}
\end{table}

\begin{table}[htbp]
\centering
\caption{Per-Block Statistics for Legal Term Density (Part 2)}
\label{tab:legal_term_density_per_block_part2}
\small
\begin{tabular}{|m{2.2cm}|m{1.8cm}|m{1.8cm}|m{1.8cm}|c|c|}
\hline
\textbf{Block} & \textbf{Pearson r (p)} & \textbf{Spearman r (p)} & \textbf{Kendall r (p)} & \textbf{Lin CCC} & \textbf{MAE} \\
\hline
Plaintiff's Claims & -0.138 ($1.83 \times 10^{-3}$) & -0.109 (0.01) & -0.084 (0.01) & -0.032 & 0.491 \\
Plaintiff's Arguments & -0.199 ($6.35 \times 10^{-6}$) & -0.134 ($2.45 \times 10^{-3}$) & -0.101 ($2.34 \times 10^{-3}$) & -0.045 & 0.479 \\
Defendant's Arguments & -0.203 ($3.87 \times 10^{-6}$) & -0.133 ($2.60 \times 10^{-3}$) & -0.099 ($2.46 \times 10^{-3}$) & -0.057 & 0.445 \\
Court Evidence Eval. & -0.105 (0.02) & -0.097 (0.03) & -0.074 (0.02) & -0.057 & 0.329 \\
Judge's Reasoning & -0.367 ($1.33 \times 10^{-17}$) & -0.319 ($1.68 \times 10^{-13}$) & -0.241 ($1.05 \times 10^{-13}$) & -0.188 & 0.345 \\
Legal Norms & -0.305 ($2.23 \times 10^{-12}$) & -0.270 ($6.05 \times 10^{-10}$) & -0.202 ($4.89 \times 10^{-10}$) & -0.143 & 0.339 \\
Court Decision & -0.162 ($2.50 \times 10^{-4}$) & -0.053 (0.23) & -0.043 (0.23) & -0.024 & 0.544 \\
\hline
\end{tabular}
\end{table}

\begin{table}[htbp]
\centering
\caption{Per-Block Statistics for LLM Evaluation Score (Part 1)}
\label{tab:llm_evaluation_score_per_block_part1}
\small
\begin{tabular}{|m{2.2cm}|c|c|c|c|c|}
\hline
\textbf{Block} & \textbf{\# Eval.} & \textbf{Mean} & \textbf{Var.} & \textbf{Std. Dev.} & \textbf{Avg. Experts} \\
\hline
Plaintiff's Claims & 995 & 0.866 & 0.054 & 0.232 & 2.028 \\
Plaintiff's Arguments & 995 & 0.854 & 0.055 & 0.235 & 2.024 \\
Defendant's Arguments & 995 & 0.854 & 0.057 & 0.238 & 2.018 \\
Court Evidence Eval. & 995 & 0.839 & 0.057 & 0.238 & 2.010 \\
Judge's Reasoning & 995 & 0.832 & 0.055 & 0.234 & 2.008 \\
Legal Norms & 995 & 0.829 & 0.057 & 0.238 & 2.014 \\
Court Decision & 995 & 0.871 & 0.053 & 0.229 & 2.018 \\
\hline
\end{tabular}
\end{table}

\begin{table}[htbp]
\centering
\caption{Per-Block Statistics for LLM Evaluation Score (Part 2)}
\label{tab:llm_evaluation_score_per_block_part2}
\small
\begin{tabular}{|m{2.2cm}|m{1.8cm}|m{1.8cm}|m{1.8cm}|c|c|}
\hline
\textbf{Block} & \textbf{Pearson r (p)} & \textbf{Spearman r (p)} & \textbf{Kendall r (p)} & \textbf{Lin CCC} & \textbf{MAE} \\
\hline
Plaintiff's Claims & 0.238 ($5.66 \times 10^{-8}$) & 0.123 ($5.68 \times 10^{-3}$) & 0.111 ($5.57 \times 10^{-3}$) & 0.235 & 0.180 \\
Plaintiff's Arguments & 0.286 ($4.76 \times 10^{-11}$) & 0.140 ($1.59 \times 10^{-3}$) & 0.124 ($1.50 \times 10^{-3}$) & 0.282 & 0.194 \\
Defendant's Arguments & 0.222 ($4.52 \times 10^{-7}$) & 0.102 (0.02) & 0.090 (0.02) & 0.209 & 0.223 \\
Court Evidence Eval. & 0.243 ($2.82 \times 10^{-8}$) & 0.189 ($1.85 \times 10^{-5}$) & 0.161 ($1.91 \times 10^{-5}$) & 0.179 & 0.292 \\
Judge's Reasoning & 0.265 ($1.23 \times 10^{-9}$) & 0.191 ($1.53 \times 10^{-5}$) & 0.164 ($1.43 \times 10^{-5}$) & 0.220 & 0.251 \\
Legal Norms & 0.239 ($4.99 \times 10^{-8}$) & 0.166 ($1.72 \times 10^{-4}$) & 0.144 ($1.55 \times 10^{-4}$) & 0.191 & 0.261 \\
Court Decision & 0.283 ($8.22 \times 10^{-11}$) & 0.144 ($1.16 \times 10^{-3}$) & 0.137 ($1.18 \times 10^{-3}$) & 0.243 & 0.142 \\
\hline
\end{tabular}
\end{table}

\begin{table}[htbp]
\centering
\caption{Document-Level Metrics Statistics (Part 1)}
\label{tab:doc_stats_part1}
\small
\begin{tabular}{|m{2.2cm}|c|c|c|c|}
\hline
\textbf{Metric} & \textbf{\# Eval.} & \textbf{Mean} & \textbf{Var.} & \textbf{Std. Dev.} \\
\hline
Coverage Ratio & 995 & 0.685 & 0.026 & 0.163 \\
Redundancy Penalty & 995 & 0.512 & 0.030 & 0.173 \\
Compression Ratio & 995 & 0.274 & 0.030 & 0.173 \\
Term Freq. Coherence & 995 & 0.720 & 0.028 & 0.168 \\
Citation Coverage & 995 & 0.556 & 0.026 & 0.162 \\
Semantic Entropy & 995 & 0.711 & 0.011 & 0.106 \\
Raw Cosine Similarity & 995 & 0.603 & 0.015 & 0.123 \\
Block Order Consistency & 995 & 0.680 & 0.013 & 0.115 \\
Monotonicity Score & 995 & 0.500 & 0.000 & 0.000 \\
Block Completeness & 995 & 0.685 & 0.026 & 0.163 \\
Keyword-Based Pseudo-F1 & 995 & 0.634 & 0.016 & 0.125 \\
\hline
\end{tabular}
\end{table}

\begin{table}[htbp]
\centering
\caption{Document-Level Metrics Statistics (Part 2)}
\label{tab:doc_stats_part2}
\small
\begin{tabular}{|m{2.2cm}|m{1.8cm}|m{1.8cm}|m{1.8cm}|c|c|}
\hline
\textbf{Metric} & \textbf{Pearson r (p)} & \textbf{Spearman r (p)} & \textbf{Kendall r (p)} & \textbf{Lin CCC} & \textbf{MAE} \\
\hline
Coverage Ratio & 0.513 ($1.50 \times 10^{-35}$) & 0.353 ($2.18 \times 10^{-16}$) & 0.253 ($3.17 \times 10^{-16}$) & 0.443 & 0.139 \\
Redundancy Penalty & -0.404 ($2.34 \times 10^{-21}$) & -0.389 ($8.02 \times 10^{-20}$) & -0.272 ($3.45 \times 10^{-19}$) & -0.175 & 0.309 \\
Compression Ratio & 0.412 ($2.89 \times 10^{-22}$) & 0.390 ($6.48 \times 10^{-20}$) & 0.276 ($1.09 \times 10^{-19}$) & 0.069 & 0.497 \\
Term Freq. Coherence & 0.540 ($8.30 \times 10^{-40}$) & 0.320 ($1.33 \times 10^{-13}$) & 0.222 ($3.16 \times 10^{-13}$) & 0.512 & 0.127 \\
Citation Coverage & 0.378 ($9.57 \times 10^{-19}$) & 0.235 ($7.69 \times 10^{-8}$) & 0.165 ($6.96 \times 10^{-8}$) & 0.194 & 0.232 \\
Semantic Entropy & 0.444 ($4.79 \times 10^{-26}$) & 0.364 ($1.95 \times 10^{-17}$) & 0.255 ($4.84 \times 10^{-17}$) & 0.381 & 0.119 \\
Raw Cosine Similarity & 0.207 ($2.43 \times 10^{-6}$) & 0.082 (0.06) & 0.055 (0.07) & 0.115 & 0.203 \\
Block Order Consistency & 0.245 ($2.30 \times 10^{-8}$) & 0.078 (0.08) & 0.055 (0.07) & 0.191 & 0.149 \\
Monotonicity Score & -- & -- & -- & -- & 0.286 \\
Block Completeness & 0.513 ($1.50 \times 10^{-35}$) & 0.353 ($2.18 \times 10^{-16}$) & 0.253 ($3.17 \times 10^{-16}$) & 0.443 & 0.139 \\
Keyword-Based Pseudo-F1 & 0.308 ($1.29 \times 10^{-12}$) & 0.030 (0.49) & 0.019 (0.53) & 0.203 & 0.173 \\
\hline
\end{tabular}
\end{table}

\begin{table}[htbp]
\centering
\caption{Per-Block Expert Statistics (Part 1)}
\label{tab:per_block_expert_part1}
\small
\begin{tabular}{|m{2.2cm}|c|c|c|c|c|}
\hline
\textbf{Block} & \textbf{\# Eval.} & \textbf{Mean} & \textbf{Var.} & \textbf{Std. Dev.} & \textbf{Avg. Experts} \\
\hline
Plaintiff's Claims & 1028 & 0.872 & 0.046 & 0.214 & 2.263 \\
Plaintiff's Arguments & 1030 & 0.824 & 0.052 & 0.228 & 2.258 \\
Defendant's Arguments & 1025 & 0.775 & 0.059 & 0.243 & 2.251 \\
Court Evidence Eval. & 1019 & 0.624 & 0.072 & 0.268 & 2.244 \\
Judge's Reasoning & 1020 & 0.676 & 0.070 & 0.265 & 2.235 \\
Legal Norms & 1023 & 0.655 & 0.062 & 0.249 & 2.247 \\
Court Decision & 1023 & 0.950 & 0.028 & 0.166 & 2.249 \\
\hline
\end{tabular}
\end{table}

\begin{table}[htbp]
\centering
\caption{Per-Block Expert Statistics (Part 2)}
\label{tab:per_block_expert_part2}
\small
\begin{tabular}{|m{2.2cm}|c|c|c|c|c|c|}
\hline
\textbf{Block} & \textbf{1 Expert} & \textbf{2 Experts} & \textbf{3 Experts} & \textbf{4 Experts} & \textbf{5 Experts}  & \textbf{ICC(2,k)} \\
\midrule
Plaintiff's claims & 96 & 317 & 80 & 12 & 2 & 0.812 \\
Plaintiff's arguments & 97 & 319 & 79 & 12 & 2 & 0.757\\
Defendant's arguments & 99 & 316 & 79 & 13 & 1 & 0.820\\
Court evaluation of evidence & 101 & 315 & 77 & 13 & 1 & 0.855  \\
Judge's reasoning steps & 99 & 320 & 76 & 12 & 1 & 0.846 \\
Applicable legal norms & 100 & 316 & 78 & 13 & 1  & 0.796 \\
Court decision & 98 & 317 & 78 & 13 & 1 & 0.695 \\
\hline
\end{tabular}
\end{table}

\clearpage

\end{appendices}

\end{document}